\pdfoutput=1

\documentclass[11pt]{article}

\usepackage{acl}

\usepackage{times}
\usepackage{latexsym}

\usepackage[T1]{fontenc}

\usepackage[utf8]{inputenc}

\usepackage{microtype}

\usepackage{inconsolata}

\usepackage{graphicx}
\usepackage{mdframed}       

\usepackage{amssymb}
\usepackage{amsmath}

\usepackage{graphicx}%
\usepackage{multirow}%
\usepackage{amsmath,amssymb,amsfonts}%
\usepackage{amsthm}%
\usepackage{mathrsfs}%
\usepackage[title]{appendix}%
\usepackage{xcolor}%
\usepackage{cite}%
\usepackage{textcomp}%
\usepackage{manyfoot}%
\usepackage{booktabs}%
\usepackage{algorithm}%
\usepackage{algorithmicx}%
\usepackage{algpseudocode}%
\usepackage{listings}%
\usepackage{utfsym}%

\usepackage{indentfirst}
\usepackage[utf8]{inputenc}

\usepackage{makecell}
\usepackage{subfigure}

\title{BASE-SQL: A powerful open source Text-To-SQL baseline approach}

\author{Lei Sheng\footnotemark[1] \\\makecell{  Wuhan University\\ of Technology, China} \\  \texttt{xuanfeng1992@whut.edu.cn} \\
  \And
  Shuai-Shuai Xu \\ \makecell{ University of Science and \\Technology of China, China} \\ \texttt{sa517432@mail.ustc.edu.cn} \\
  \And
  Wei Xie \\ \makecell{Independent \\researcher, China} \\ \texttt{xiedaren95@gmail.com}
}

\usepackage{hyperref}

\begin{document}
\maketitle

\footnotetext[1]{Corresponding authors.}

\begin{abstract}
The conversion of natural language into SQL language for querying databases (Text-to-SQL) has broad application prospects and has attracted widespread attention. At present, the mainstream Text-to-SQL methods are mainly divided into in-context learning (ICL) based methods and supervised fine-tuning (SFT) based methods. ICL-based methods can achieve relatively good results thanks to the use of the most advanced closed-source models. However, in real-world application scenarios, factors such as data privacy, SQL generation efficiency and cost need to be considered. SFT-based methods have certain advantages. At present, methods based on fine-tuning of open source models lack easy-to-implement and effective (cost-effective) baseline methods. We propose a pipeline-based method using open source model fine-tuning, referred to as BASE-SQL, which includes four components: Schema Linking, Candidate SQL Generate, SQL Revision and SQL Merge Revision. Experimental results show that BASE-SQL uses the open source model Qwen2.5-Coder-32B-Instruct, and achieves an accuracy of 67.47\% on the BIRD development set and 88.9\% on the Spider test set, which is significantly better than other methods using open source models, and even exceeds several methods using the GPT-4o closed-source model. At the same time, BASE-SQL is easy to implement and highly efficient (on average, only five calls to the large language model are required to generate SQL once). The code will be open sourced at \url{https://github.com/CycloneBoy/base_sql}.
\end{abstract}

\section{Introduction}
\label{sec:introduction}

Text-to-SQL task is to convert natural language into Structured Query Language (SQL) for querying database, which has received more and more attention\citep{qin_survey_2022, katsogiannis-meimarakis_survey_2023, liu_survey_2024, shi_survey_2024}. It can reduce the difficulty of interaction between users and databases, especially in the fields of data analysis, business intelligence, intelligent customer service and other broad application prospects. With the rapid development of deep learning technology, Text-to-SQL tasks have made rapid progress, from only being able to answer the domain-specific single-table questions before, to being able to solve cross-domain multi-table complex problem SQL generation. On the famous Spider\citep{yu_spider_2019} benchmark, it has achieved an accuracy rate of 90\%.

Early Text-to-SQL methods were mainly based on the sequence-to-sequence (Seq2Seq) generation method\citep{wang-etal-2020-rat, guo-etal-2019-towards, Li_Zhang_Li_Chen_2023}. The current mainstream is the method based on large language models (LLMs), which can be roughly divided into in-context learning (ICL) based methods and supervised fine-tuning (SFT) based methods\citep{liu_survey_2024}. ICL-based methods\citep{pourreza_din-sql_2023, gao_text--sql_2023, lee_mcs-sql_2024, pourreza2024chase} mainly use the most advanced closed-source LLMs(such as: GPT-4, GPT-4o, Gemini-1.5 Pro, etc.), and then use various strategies to improve state-of-the-art (SOTA) performance without considering the efficiency and cost of SQL generation. Benefiting from the powerful reasoning ability of the model, it often achieves better results. SFT-based methods use open source LLMs for fine-tuning to improve the effect of SQL generation\citep{pourreza-rafiei-2024-dts,li_codes_2024, yang_synthesizing_2024}. Due to the small size of the model used, it often performs worse than ICL-based methods.

In real-world Text-to-SQL application scenarios, the following factors will be considered: data privacy, efficiency and cost of SQL generation. Therefore, the method of SFT-based on open source models has certain advantages. CodeS\citep{li_codes_2024} uses a specially curated SQL-centric corpus for incremental pre-training, and combines it with a two-way data augmentation technique for SFT, which ultimately exceeds the methods that use a large number of closed-source models, but it relies on a large amount of data for incremental pre-training. SENSE\citep{yang_synthesizing_2024} uses data synthesized by closed-source models to fine-tune small models, demonstrating the effectiveness of its synthetic data, but it relies on closed-source models for data synthesis. DTS-SQL\citep{pourreza-rafiei-2024-dts} simplifies SQL generation by fine-tuning schema linking and SQL generation separately, but lacks detailed analysis. CHESS\citep{talaei_chess_2024} uses a pipeline method to generate SQL in multiple steps, proving the superiority of the pipeline method. MSc-SQL\citep{gorti2024msc} generates candidate SQL by fine-tuning multiple different models, and then uses a selection model to select the final results. Although these methods are comparable to the methods using the closed-source model like GPT-4, they are lower than the methods using the most advanced closed-source models like GPT-4o and Gemini 1.5 pro. At the same time, the SFT-based methods lack a high cost-effectiveness baseline method, which also hinders the development of SFT-based methods.

In order to solve the challenges faced by the open source model SFT method, we propose a new method BASE-SQL based on the pipeline method. BASE-SQL mainly consists of four components: Schema Linking, Candidate SQL Generate, SQL Revision and SQL Merge Revision. In the Schema Linking component, we only perform table-linking, not column-linking. Table-linking is obtained by fine-tuning an open source model. Candidate SQL Generate is also obtained by fine-tuning an open source model. SQL Revision uses another open source model combined with Full Schema to perform SQL error correction. Finally, SQL Merge Revision is used to generate the final SQL through two rounds of merge correction. Through detailed experiments, we analyzed the effects of different components under different parameter configurations. Finally, our method BASE-SQL only uses two 32B open source models Qwen2.5-Coder-32B-Instruct\citep{hui2024qwen2} and Qwen2.5-32B-Instruct\citep{qwen2.5}, and achieved an accuracy of 67.47\% on the BIRD\citep{li2024can} development set and 88.9\% on the Spider test set. BASE-SQL is much more efficient than other state-of-the-art methods. On average, it only needs to call the large model 5 times for one generation. In addition, the cost of reproducing BASE-SQL is also very low. There is no need to use additional data for incremental pre-training. In general, BASE-SQL is a cost-effective Text-to-SQL baseline method.

In summary, our main contributions are as follows:
\begin{enumerate}
    \item  We analyzed the advantages and disadvantages of the current SFT-based Text-to-SQL methods and proposed a new pipeline-based method BASE-SQL.

    \item Through a large number of comparative experiments, we analyzed the SQL generation effects of different components of our method under different parameter configurations, providing a certain reference for subsequent research.

    \item  We only use two 32B open source models, and on the BIRD and Spider datasets, we outperform all other methods using open source models, and even surpass several methods using GPT-4o closed source models. At the same time, it has high inference efficiency and low reproduction cost, indicating that BASE-SQL is a cost-effective baseline method. Our source code will be published at \url{https://github.com/CycloneBoy/base_sql}.
\end{enumerate}

\section{Related Work}
\label{sec:related_works}
Early Text-to-SQL methods are mainly based on the Seq2Seq method. The natural language and database schema are semantically encoded through the encoder, and then decoded through the decoder to generate the corresponding SQL statement\citep{zhong2017seq2sqlgeneratingstructuredqueries}. Since the SQL language has certain grammatical rules, it is easy to make errors when generating SQL models directly through the decoder. On this basis, slot filling methods are based on SQL grammatical rules \citep{wang-etal-2020-rat, guo-etal-2019-towards, yu-etal-2018-syntaxsqlnet} and sketch-based methods \citep{lyu2020hybridrankingnetworktexttosql, yu-etal-2018-typesql} are proposed. For example, \citep{Li_Zhang_Li_Chen_2023} proposes a ranking-augmented encoder to alleviate the workload of schema linking and a skeleton-aware decoder to implicitly guide the SQL generation of skeletons. These methods are limited by the capabilities of the basic model, the model generalization ability is not strong, and a large amount of data is required for fine-tuning for databases in different domains.

With the rapid development from pre-trained language model to large language model, LLM-based Text-to-SQL methods have become mainstream. DIN-SQL\citep{pourreza_din-sql_2023} decomposes the generation problem into multiple sub-problems  and uses GPT-4 for ICL, and finally defeats a large number of fine-tuned models. DAIL-SQL\citep{gao_text--sql_2023} explores the advantages and disadvantages of different schema representations, example selections, and example organizations through a large number of comparative experiments, providing effective suggestions for subsequent research. MCS-SQL\citep{lee_mcs-sql_2024} uses different prompts to explore a broader search space to generate multiple candidate SQLs, then filters them based on confidence scores, and finally uses multiple-choice selections to obtain the final results. \citep{wang_mac-sql_2024,cen_sqlfixagent_2024} use a multi-agent collaboration framework and use different agents to complete specific subtasks and finally complete SQL generation. CHASE-SQL\citep{pourreza2024chase} found that the widely used self-consistency strategy could not effectively select the best SQL from a large amount candidate SQL pools, so it proposed a sorting strategy based on pairwise comparisons and combined three innovative strategies to generate diverse and high-quality candidate SQL, achieving SOTA on the BIRD dataset. Since the ICL-based method does not require additional computing resources for model fine-tuning and has high flexibility and generalization, it is currently a widely used method.

Due to issues such as data privacy and efficiency, methods based on SFT have certain advantages. \citep{pourreza-rafiei-2024-dts,li_codes_2024, gorti2024msc} selected open source LLM for fine-tuning and achieved good results, surpassing many methods that use proprietary closed source LLM. CodeS\citep{li_codes_2024} combines schema linking, data enhancement, and pre-training methods to achieve SOTA results on multiple datasets, and open-sources pre-trained models from 1B to 15B. SENSE\citep{yang_synthesizing_2024} uses strong LLM to synthesize weakly supervised data to train small LLM. The experimental results show the effectiveness of its synthetic data.  MSc-SQL\citep{gorti2024msc} first filters the schema by schema linking, then uses multiple small LLM to fine-tune to generate a variety of candidate SQLs, and finally uses a selector to select the final SQL. CHESS\citep{talaei_chess_2024} and XiYan-SQL\citep{xiyansql} combine the advantages and disadvantages of ICL and model fine-tuning, based on a pipeline approach, including entity and context retrieval, schema linking, and fine-tuned LLM for SQL generation and SQL selection, proving the effectiveness of the combination of ICL and SFT methods.


\section{Methodology}
\label{sec:methods}
\begin{figure*}[t]
	\begin{center}
		  \includegraphics[width=1\textwidth]{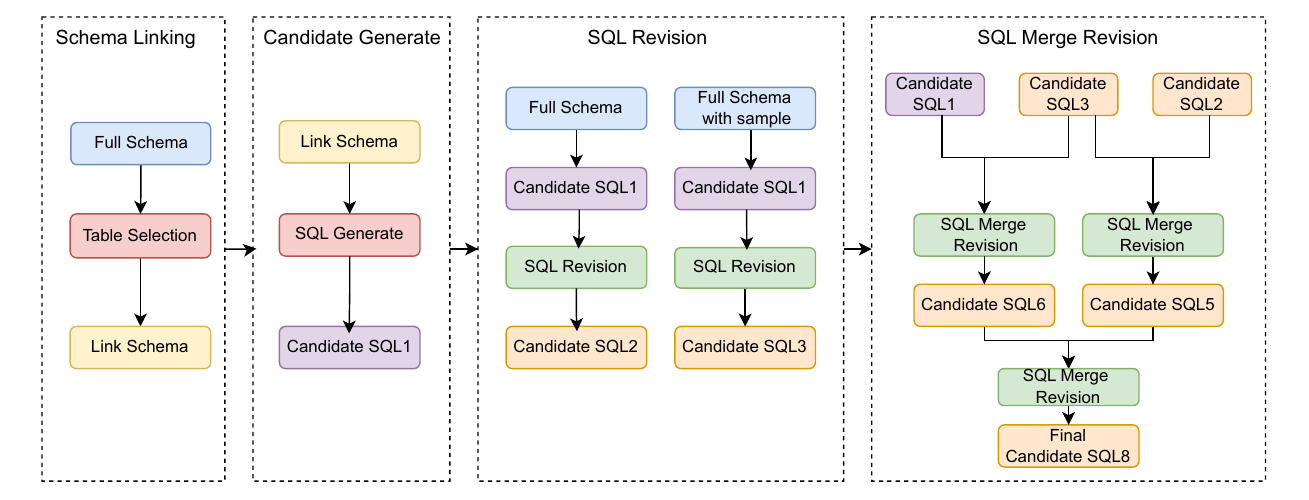}
		  \caption{Overview of the proposed BASE-SQL framework, which consists of four components:  1) Schema linking: Identify the most relevant tables through a fine-tuned model; 2) Candidate Generation: Generate candidate SQL1 through a fine-tuned model; 3) SQL Revision: Use the all table schema, candidate SQL1 and its execution results to perform SQL correction and generate candidate SQL2 and candidate SQL3; 4) SQL Merge Revision: Use candidate SQL1, candidate SQL2 and candidate SQL3 to perform combined correction to generate the final SQL.}
		  \label{figure:base_sql} 
	\end{center}
\end{figure*}

The framework of BASE-SQL we proposed is shown in Figure \ref{figure:base_sql}, which includes four components: Schema Linking, Candidate SQL Generate, SQL Revision and SQL Merge Revision.

\subsection{Schema Representation}
\label{subsec:schema_Representation}

In order for LLM to generate the corresponding SQL, the table structure information of the database needs to be included in the prompt context so that LLM can fully understand the database schema. There are many ways to represent database table structure information\citep{gao_text--sql_2023}, including code structure representation\citep{nan-etal-2023-enhancing}, Alpaca SFT Prompt representation\citep{lee_mcs-sql_2024}, Chess representation\citep{talaei_chess_2024}, M-Schema representation\citep{xiyansql}, etc. More and more representations include sample values of columns, which allows LLM to understand the data in the table. The M-Schema representation proposed by XiYan-SQL\citep{xiyansql} recently shows the column name, data type, column descriptor, primary key information and sample value of each column in the database, which performs better than the code structure representation. Although each column contains the corresponding sample value, our experiment found that adding a three-column table result of random query of the current table after each table can further improve LLM's understanding of the database table structure. Figure \ref{figure:schema_representation} shows different representation methods.

\subsection{Schema Linking}
\label{subsec:schema_linking}
Schema linking is to identify the database tables and column information necessary to convert natural language into SQL, and filter out the noise information brought by redundant table columns. It can significantly improve the accuracy of SQL generation. Currently, it mainly includes three methods: database entity linking, table linking, and column linking\citep{liu_survey_2024}. Among them, the column linking is the most difficult, because missing any column will cause the subsequent SQL generation to fail. In order to improve the recall rate of column linking, many methods usually need to call LLM multiple times in this part, resulting in low efficiency of SQL generation. For example, XiYan-SQL\citep{xiyansql} and CHESS\citep{talaei_chess_2024} both use Column Selector to filter irrelevant column. It needs to judge whether each column in the database and the user question needs to be retained.

Like MSc-SQL\citep{gorti2024msc}, we only use table linking to identify the table names related to the question through a fine-tuned model. We do not use column linking for the following three reasons: 1) The open source model we use is relatively small, and the low recognition accuracy of column linking will reduce the accuracy of subsequent SQL generation; 2) We found through experiments that for our model, even if the column linking is completely accurate, the improvement in SQL generation accuracy is not significant (see experimental results from Table \ref{tab:result_different_schema_linking_method}). 3) In order to take into account the efficiency of overall SQL generation, reduce the number of calls to the LLM.

\subsection{Candidate Generation}
\label{subsec:candidate_generation}
We use the powerful open source model Qwen2.5-Coder-32B-Instruct\citep{hui2024qwen2} as the base model. It has achieved SOTA performance on multiple code-related benchmarks. Inspired by MSc-SQL\citep{gorti2024msc}, in order to simulate the redundant noise caused by identifying redundant tables in the Schema Linking process, we randomly selected 10\% of the samples based on the ground-truth table in the training samples according to the results of Schema Linking to add noise (randomly add 1 to 2 additional tables after excluding the ground-truth table from the selected samples).

Thanks to the powerful Qwen2.5-Coder\citep{hui2024qwen2} pre-trained model, we found in the experiment that the model can achieve high accuracy by fine-tuning only a small number of samples (1k-4k) using LoRA\citep{hu2021lora} fine-tuning (see experimental results from Figure \ref{figure:sft_step}).

\subsection{SQL Revision}
\label{subsec:sql_revision}
The generated candidate SQL may contain logic and syntax errors. Most methods\citep{talaei_chess_2024,pourreza2024chase,xiyansql} use multiple rounds of SQL correction to improve the accuracy of SQL. They usually input the table structure after Schema Linking, the generated candidate SQL and its execution results into LLM, and use LLM to correct potential errors in SQL. Through experiments, we found that using the table structure information before Schema Linking can further improve the accuracy of correction, because some table recognition errors will exist after Schema Linking. We use the M-Schema representation and M-Schema With Sample representation of all tables to perform secondary independent corrections and generate candidate SQL2 and candidate SQL3 respectively.

\subsection{SQL Merge Revision}
\label{subsec:sql_merge_revision}
The process of LLM generation is a probabilistic sampling process. It is generally difficult to guarantee that the best result is generated once. Therefore, many methods\citep{xiyansql,pourreza2024chase,gorti2024msc,lee_mcs-sql_2024} will generate multiple candidate SQLs and then select the final SQL in different ways. Typical methods include: self-consistency, re-ranking method, selection method. Self-consistency has certain limitations and cannot guarantee that the most consistent result is the correct result. Recently, model selection methods have received attention and achieved significant performance improvements. At present, the selection model needs to be fine-tuned with the corresponding data. The effect of the un-fine-tuned selection model is not ideal (see experimental results from Table \ref{tab:result_different_sql_selection}). However, there is currently a lack of corresponding open source datasets.

Therefore, we propose a merge correction method, which combines the characteristics of correction and selection and generates the final SQL through multiple merge corrections. SQL merge correction is only required when the execution results of two candidate SQLs are inconsistent. Otherwise, we select the first candidate SQL. First, the table columns that appear in the two candidate SQLs are used as the table structure information for merge correction. Then, the two candidate SQLs and their corresponding execution results are listed, and LLM generates a new SQL instead of selecting one of them as the candidate SQL. The candidate SQL2, SQL3 and candidate SQL1 before correction obtained by the SQL correction step are merged and corrected three times to obtain the final candidate SQL.

\section{Experiments}
\label{sec:experiments}

\subsection{Experiments Setting}
\label{subsec:experiments_setting}

\textbf{Datasets} We evaluate our method on two widely used Text-to-SQL datasets: Spider\citep{yu_spider_2019} and BIRD\citep{li_can_2023}. The detailed information of the two datasets is shown in Appendix \ref{sec:appendix_experiment_dataset}.

\textbf{Metrics} Execution accuracy(EX): The accuracy of the predicted SQL is evaluated by comparing the results of the predicted SQL execution with the results of the GOLD-SQL execution in the validation database. It is the official metric used by Spider and BIRD as their leaderboards, and is also the main metric for our experimental comparison.

\textbf{Implementation details} See Appendix \ref{sec:appendix_implementation_details} for implementation details.

\subsection{Main Results}
\label{subsec:main_result}

 \begin{table*}[htbp]
    \centering
	\begin{tabular*}{\textwidth}{llllcccc}    
		\toprule
        \text{Method}  & \text{Type}  & \text{Model}   & \text{Size}  & \text{EX(Dev)}  & \text{EX(Test)}\\
		\hline

        \text{CHASE-SQL\citep{pourreza2024chase}}  & \text{ICL}  & \text{Gemini 1.5 pro}  & \text{UNK} & \textbf{74.46} & \text{74.79}  \\
		\text{XiYan-SQL\citep{xiyansql}}  & \text{\makecell{ICL+\\SFT}}  & \text{GPT-4o} &\text{UNK} & \text{73.34} & \textbf{75.63}  \\
        \text{CHESS\citep{talaei_chess_2024}}  & \text{ICL}  & \text{Gemini 1.5 pro}  &\text{UNK} & \text{68.31} & \text{71.10}  \\
        \text{Distillery\citep{maamari_death_2024}}  & \text{SFT}  & \text{GPT-4o}  &\text{UNK} & \text{67.21} & \text{71.83}  \\
        \text{RSL-SQL\citep{cao2024rsl}}  & \text{ICL}  & \text{GPT-4o}  &\text{UNK} & \text{67.21} & \text{-}  \\
        \text{E-SQL\citep{caferoglu_e-sql_2024}}  & \text{ICL}  & \text{GPT-4o}  &\text{UNK} & \text{65.58} & \text{66.29}  \\
        \text{MCS-SQL\citep{lee_mcs-sql_2024}}  & \text{ICL}  & \text{GPT-4}  &\text{UNK} & \text{63.40} & \text{65.50}  \\
        \text{PTD-SQL\citep{luo-etal-2024-ptd}}  & \text{ICL}  & \text{GPT-4}  &\text{UNK} & \text{57.0} & \text{-}  \\
        \text{TA-SQL\citep{qu_before_2024}}  & \text{ICL}  & \text{GPT-4}  &\text{UNK} & \text{56.19} & \text{59.14}  \\
        \text{DIN-SQL\citep{pourreza_din-sql_2023}}  & \text{ICL}  & \text{GPT-4}  &\text{UNK} & \text{50.72} & \text{55.90}  \\
         \midrule
         \text{MSc-SQL\citep{gorti2024msc}}  & \text{SFT}  & \text{Mistral-v0.3,...}  &\text{9B} & \text{65.60} & \text{-}  \\
         \text{CHESS\citep{talaei_chess_2024}}  & \text{\makecell{ICL+\\SFT}}  & \text{Llama-3}  &\text{70B} & \text{61.50} & \text{-}  \\
         \text{Distillery\citep{maamari_death_2024}}  & \text{ICL}  & \text{Llama-3.1}  &\text{405B} & \text{59.18} & \text{-}  \\
         \text{CodeS\citep{li_codes_2024}}  & \text{SFT}  & \text{StarCoder}  &\text{15B} & \text{58.47} & \text{60.37}  \\
         \text{CodeS\citep{li_codes_2024}}  & \text{SFT}  & \text{StarCoder}  &\text{7B} & \text{57.17} & \text{59.25}  \\
         \text{DTS-SQL\citep{pourreza-rafiei-2024-dts}}  & \text{SFT}  & \text{DeepSeek-Coder}  &\text{7B} & \text{55.80} &  \text{-}  \\
         \text{SENSE\citep{yang_synthesizing_2024}}  & \text{SFT}  & \text{CodeLlama}  &\text{13B} & \text{55.48} &  \text{63.39}  \\
         \text{SENSE\citep{yang_synthesizing_2024}}  & \text{SFT}  & \text{CodeLlama}  &\text{7B} & \text{51.80} &  \text{59.30}  \\
         \midrule
         \text{Qwen2.5-Coder\citep{hui2024qwen2}}  & \text{ICL}  & \text{Qwen2.5-Coder}  &\text{32B} & \text{58.40} & \text{-}  \\
         \textbf{BASE-SQL(Ours)}  & \text{SFT}  & \text{Qwen2.5-Coder}  &\text{32B} & \text{\underline{67.47}}(\text{$\uparrow$9.07}) & \text{-}  \\
         \text{DeepSeek-Coder\citep{deepseek-coder}}  & \text{ICL}  & \text{DeepSeek-Coder}  &\text{33B} & \text{49.54} & \text{-}  \\
         \textbf{BASE-SQL(Ours)}  & \text{SFT}  & \text{DeepSeek-Coder}  &\text{33B} & \text{\underline{65.38}}(\text{$\uparrow$15.84}) & \text{-}  \\
         \text{Qwen2.5-Coder\citep{hui2024qwen2}}  & \text{ICL}  & \text{Qwen2.5-Coder}  &\text{14B} & \text{56.90} & \text{-}  \\
         \textbf{BASE-SQL(Ours)}  & \text{SFT}  & \text{Qwen2.5-Coder}  &\text{14B} & \text{\underline{63.82}}(\text{$\uparrow$6.92}) & \text{-}  \\
		\bottomrule
	\end{tabular*}
	\caption{Performance Comparison of different Text-to-SQL methods on BIRD dev and test dataset. The Type column represents the type of method, including: In-context learning(ICL), supervised fine-tuning(SFT), and ICL+SFT. The Size column represents the number of parameters of the model, where UNK represents unknown size and B represents billions. EX represents execution accuracy. The underline represents the best result of our method.}
	\label{tab:result_main_bird}
	\footnotetext{}
\end{table*}

\begin{table*}[htbp]
    \centering
	\begin{tabular}{lcc|cc|cc}    
		\toprule
        \text{Method} & \text{\makecell{BIRD\\EX(Dev)}}&  ${\Delta}$\text{EX} & \text{\makecell{BIRD\\EX(Dev)}}&  ${\Delta}$\text{EX} & \text{\makecell{Spider\\EX(Test)}} & ${\Delta}$\text{EX}\\
		\hline
        \text{model}  & \makecell{Qwen2.5-\\Coder(32B)} & \text{-} & \makecell{DeepSeek-\\Coder(33B)} & \text{-}   & \makecell{Qwen2.5-\\Coder(32B)} & \text{-} \\
        \hline
        \text{BASELINE}  & \text{56.84} & \text{-}  & \text{45.37} & \text{-}   & \text{82.16} & \text{-} \\
        \text{+ M-Schema} & \text{58.21} & \text{+1.37} & \text{49.54} & \text{+4.17} & \text{83.42} & \text{+1.26} \\
        \text{+ Data Samples} & \text{59.71} & \text{+1.50} & \text{44.26} & \text{-5.28} & \text{85.61} & \text{+2.19} \\
        \text{+ Schema Linking} & \text{61.54} & \text{+1.83} & \text{50.59} & \text{+6.33} & \text{85.10} & \text{-0.51}   \\
        \text{+ Supervised Fine-tuning} & \text{64.21} & \text{+2.67} & \text{59.06} & \text{+8.47}  & \text{87.47} & \text{+2.37}  \\
        \text{+ SQL Revision} & \text{66.04} & \text{+1.83} & \text{64.41} & \text{+5.35}  & \text{88.12} & \text{+0.65}  \\
        \text{+ SQL Merge Revision} & \textbf{67.47} & \text{+1.43} & \textbf{65.38} & \text{+0.97} & \textbf{88.87} & \text{+0.75}  \\
		\bottomrule
	\end{tabular}
	\caption{Ablations of our method on BIRD development dataset and Spider test dataset.}
	\label{tab:result_ablation_bird}
	\footnotetext{}
\end{table*}

\textbf{BIRD results} We compared with the most advanced methods, and the results are shown in Table \ref{tab:result_main_bird}. Among the methods that use open source LLMs (the middle and lower results), our method (use Qwen2.5-Coder-32B-Instruct) achieves the best on the development set, 1.87 points higher than the second place MSc-SQL\citep{gorti2024msc} and 9.07 points higher than the original Qwen2.5-Coder-32B-Instruct. Compared with the methods that use closed source LLMs (the top part), our method is competitive, surpassing four methods that use GPT-4 and three methods that use GPT-4o on the development set. After BASE-SQL adopts the 33B DeepSeek-Coder basic model, the performance is improved by 15.84\%; even if the 14B Qwen2.5-Coder-32B-Instruct is adopted, the performance can be improved by 6.92\%, which shows the effectiveness of BASE-SQL.

Our method only uses two open source LLMs with 32B parameters, while CHASE-SQL, XiYan-SQL, and CHESS uses the most advanced closed source LLMs and generate a large number of candidate SQLs, and then filter out the final SQL through different strategies. Overall, the generation cost is expensive and the efficiency is low. CHASE-SQL\citep{pourreza2024chase} uses three chain-of-thought prompting techniques to generate 21 candidate SQLs, and then combines a fine-tuned binary selection model to filter out the final SQL. XiYan-SQL\citep{xiyansql} uses additional data for fine-tuning and combines the ICL method to generate 5 candidate SQLs, and finally uses a fine-tuned binary selector to filter out the final SQL. CHESS\citep{talaei_chess_2024} generates 20 candidate SQLs and 10 test cases to filter out the final SQL. Among them, XiYan-SQL and CHESS use column filtering for Schema Linking, which is very inefficient. It needs to perform binary classification on each column of the database and the user question to determine whether the column is related to the user question, thereby filtering out columns that are not related to the user question. Overall, our method has achieved a certain balance in performance, cost, and efficiency.


\textbf{Spider results}: Table \ref{tab:result_main_spider} shows the results of our method on the Spider data set. Our method achieves EX of 86.9\% on the development set and 88.9\% on the test set, surpassing all methods using open source LLMs and even surpassing CHASE-SQL\citep{pourreza2024chase} and CHESS\citep{talaei_chess_2024} on the test set. It is only 0.8\% lower than the SOTA methods XiYan-SQL\citep{xiyansql} and MCS-SQL\citep{lee_mcs-sql_2024}. MCS-SQL uses the ICL method to generate up to 20 candidate SQLs, and then uses the Multiple-Choice Selection method to select the best SQL, while our method only generates 3 candidate SQLs. Among them, the 14B version of our method BASE-SQL is extremely competitive, only 0.1\% lower than the 32B version on the development set and 1.0\% lower on the test set. The experimental results show that BASE-SQL has strong generalization properties.

\subsection{Ablation Studies}
\label{subsec:ablation_studies}

\begin{table*}[htbp]
    \centering
	\begin{tabular}{l|cc|ccc|cc}    
		\toprule
        \text{Method} & \text{Accuracy} & \text{Recall} & \text{\makecell{Avg\\Precision}} & \text{\makecell{Avg\\Recall}} & \text{\makecell{Avg\\F1}}  & \text{EX}&  \text{EX(SFT)}\\
        \midrule
        \text{Full Schema} & \text{1} &\text{100} &\text{29.72} &\text{100} &\text{45.82} &\text{59.71} &\text{62.19} \\
        \text{Gold Table Schema} & \text{100} &\text{100} &\text{100} &\text{100} &\text{100} & \text{64.02} & \text{67.28} \\
        \text{Gold Column Schema} & \text{100} &\text{100} &\text{100} &\text{100} &\text{100} &\text{63.89} & \text{68.32} \\
        \midrule
        \textbf{Our Schema Linking}  & \textbf{82.39} & \text{89.83} & \textbf{94.83} & \text{95.51} & \textbf{95.17} & \textbf{61.54} & \textbf{64.21} \\
        \text{Schema Linking one}  & \text{64.34} & \text{95.31} & \text{85.45} & \text{97.95} & \text{91.27} & \text{59.84} & \text{63.75} \\
        \text{Schema Linking two}  & \text{36.05} & \textbf{97.00} & \text{70.09} & \textbf{98.69} & \text{81.97} & \text{59.39} & \text{63.23} \\
		\bottomrule
	\end{tabular}
	\caption{Experimental results using different schema linking methods. Accuracy indicates the accuracy of a predicted result that is completely consistent with the Gold table. Recall indicates the accuracy of a predicted result that includes all current Gold tables. Avg is the abbreviation for average, which calculates the precision, recall, and F1 of each predicted result and Gold table, and finally calculates the average of all three metrics. EX represents execution accuracy on the BIRD development set. SFT represents supervised fine-tuning.}
	\label{tab:result_different_schema_linking_method}
	\footnotetext{}
\end{table*}

We conducted ablation experiments on the BIRD development set and the Spider test set to analyze the impact of different components on the overall SQL generation. The experimental results are shown in Table \ref{tab:result_ablation_bird}. Our baseline setting uses the entire schema in code representation. We take the BIRD development set as an example for analysis. After replacing code representation schema with M-Schema, the performance improved by 1.37\%, demonstrating the superiority of M-Schema. Then we adopted M-Schema with sample representation, and the performance further improved by 1.5\%, indicating that adding table content can further help LLM understand the table structure information from a global perspective, although the description of each column in M-Schema already contains several sample values of the current column. After adding Schema-linking, the performance improved by 1.83\%, indicating that Schema-linking can reduce the noise in schema. After performing supervised fine-tuning, the largest performance improvement of 2.67\% was obtained, indicating the importance of supervised fine-tuning for LLM to perform downstream tasks. After SQL revision, the performance further improved by 1.83\%, indicating that LLM can correct errors in some candidate SQLs by using candidate SQLs and their execution results. Finally, SQL merge revision were performed, and the performance was improved by 1.43\%, indicating that LLM can use multiple candidate SQLs and corresponding results to further improve the accuracy of SQL generation using SQL generated in different ways. 

The performance improvement in the Spider test set was similar, except that the performance decreased by 0.51\% after adding schema linking, indicating that inaccurate schema linking will cause subsequent SQL performance to degrade\citep{maamari_death_2024}, so we use entire schema for SQL revision to reduce the impact of performance loss caused by inaccurate schema linking.

\subsection{Analysis}
\label{subsec:analysis}

\subsection{Impacts of Schema Linking}
\label{subsec:impacts_of_schema_linking}

In order to analyze the impact of Schema Linking on SQL generation, we conducted experiments on the BIRD development set. The experimental results are shown in Table \ref{tab:result_different_schema_linking_method}. The Gold Table Schema method uses all gold Tables as schemas, the Gold Column Schema method uses all gold columns as schemas, our Schema Linking method uses only a single model for fine-tuning, the Schema Linking one method uses two different models for voting, and the Schema Linking two method uses five different models for voting. By comparing the results of Gold Table Schema and Gold Column Schema, we found that their EX metric are not much different, especially the EX after fine-tuning is only 1.04\% different. Therefore, we comprehensively consider the performance and the efficiency of SQL generation, and only perform table linking instead of column linking.

By comparing the following three methods, although the Schema Linking one method and the Schema Linking two method achieve more than 95\% recall through multi-model voting, the EX is lower than that of our Schema Linking method. This shows that Schema Linking cannot focus too much on the Recall metric, and the accuracy of table prediction also needs to be considered. Our Schema linking method achieved a certain balance between precision and recall through a single model approach. After fine-tuning, the EX reached 64.21\%, which is only 3.07\% lower than the upper limit of Schema linking (using the Gold Table Schema method).

\subsection{Impacts of Different SQL Revision Methods}

We compared and analyzed different SQL revision methods, and the experimental results are shown in Table \ref{tab:result_different_revision}. First, we used three different models and four different schemas for comparison under the same pre-revison result (EX is 64.15). The experimental results are shown in columns 2, 3, and 4 in the table. When Qwen2.5-32B-Instruct is used as the SQL revision model and Full Schema, the SQL revision effect is the best. We speculate that this is because the pre-revision result is obtained by fine-tuning the same Coder model, and the model capabilities are not much different, so it is not possible to find errors well. The general Qwen2.5-32B-Instruct is used for SQL revision, which can better find SQL errors and correct them. Then, under the same SQL revision model (Qwen2.5-32B-Instruct), we used three different pre-revison results and four different Schemas for comparison. The experimental results are shown in columns 5, 6, and 7 in the table. The experimental results show that the better the pre-revison result, the better the effect after SQL revision. The Full Schema has a better effect than the schema after Schema Linking. This shows that the Full Schema post-revision model can correct some samples that were previously incorrectly identified by Schema Linking.

\subsection{Impacts of Other Analysis}
See Appendix \ref{sec:appendix_analysis} for more analysis details.

\section{Conclusion}
\label{sec:conclusion}

In this paper, we proposed the BASE-SQL framework, a pipeline-based Text-to-SQL method, which mainly consists of Schema Linking, Candidate SQL Generate, SQL Revision and SQL Merge Revision. We conducted detailed experiments and analyzed the impact of each component under different parameter configurations, providing a certain reference for subsequent research. BASE-SQL only uses two 32B open source models and achieves competitive results on BIRD and Spider datasets, even surpassing most methods using GPT-4 and GPT-4o. At the same time, BASE-SQL is easy to implement and has high SQL generation efficiency. It is a strong baseline method suitable for use in real-world application scenarios.

\section{Limitations}
\label{sec:limitations_and_future_work}

Although our method shows competitive results in various aspects, the research in this paper still has certain limitations. First, due to resource and time constraints, we cannot fully verify the effects of other open source LLMs (such as: Llama-3.3-70B-Instruct, Mistral-Large-Instruct-2411). Secondly, we only use open source LLMs for experiments, and have not verified the BASE-SQL effect of closed source LLMs (such as: GPT-4o, Gemini 1.5 pro). Then we only performed Table Linking in Schema Linking, and did not perform Column Linking. In the case of a large number of columns in the table in real-world application scenarios (hundreds or even thousands of columns), the prompt will be particularly long, which may affect the effect of SQL generation.

\bibliography{base_sql}

\appendix

\begin{table*}[htbp]
    \centering
	\begin{tabular}{l|ccc|ccc}    
		\toprule
        \text{Method} & \text{SFT Coder}&  \text{Coder} & \text{Qwen} & \text{Qwen} & \text{Qwen} & \text{Qwen} \\
        \midrule
        \text{Before SQL revision} & \text{64.15} & \text{64.15} & \text{64.15} & \text{63.62} & \text{64.15} & \text{64.21} \\
        \midrule
        \text{Schema Linking Schema} & \text{63.04} & \text{64.41} & \text{65.25} & \text{64.41} & \text{65.25} & \text{65.65}\\
        \text{Schema Linking With Sample Schema } & \text{63.43} & \text{64.21} & \text{65.19} & \text{64.67} & \text{65.19} & \text{65.45}\\
        \text{Full Schema} & \text{63.95} & \text{64.28} & \textbf{65.91} & \text{64.73} & \text{65.91} & \textbf{66.04} \\
        \text{Full With Sample Schema} & \text{63.62} & \text{64.80} & \text{65.58} & \text{64.73} & \text{65.58} & \text{65.91}\\
		\bottomrule
	\end{tabular}
	\caption{Execution accuracy(EX) of different schema representations and different revision model on the BIRD development set. SFT represents supervised fine-tuning. Coder represents Qwen2.5-Coder-32B-Instruct and Qwen represents Qwen2.5-32B-Instruct.}
	\label{tab:result_different_revision}
	\footnotetext{}
\end{table*}

\section{Experiments Setting}
\label{sec:appendix_experiment_setting}

\subsection{Dataset}
\label{sec:appendix_experiment_dataset}

Spider\citep{yu_spider_2019} is a complex cross-domain text-to-SQL benchmark that includes 10,181 questions and 5,693 independent complex SQLs in 200 databases, covering 138 different domains. It has been divided into non-overlapping training, development, and test sets. Unlike Spider, BIRD\citep{li_can_2023} focuses on real-world application scenarios, including dirty and noisy database values, external knowledge grounding between natural language questions and database values, and the efficiency of SQL execution in the context of massive databases. It includes 12,751 question and SQL pairs in 95 databases in 37 professional domains, with a total database size of 33.4GB. BIRD has been divided into public training and development sets, and its unpublished test set is used as an independent and fair verification dataset. 

\subsection{Implementation Detail}
\label{sec:appendix_implementation_details}

We uesed the open source model Qwen2.5-Coder-32B-Instruct as the basic model for Schema Linking and Candidate Generation, and used LoRA\citep{hu2021lora} to fine-tune. Fine-tuning was performed using LoRA\citep{hu2021lora},
with a LoRA rank of 32, a LoRA $\alpha$ of 16, and a dropout rate of 0.1. We fine-tune target model with accumulation batch size of 8, a learning rate of 4e\mbox{-}5, and a max context length of 4,096. All our experiments were performed using two NVIDIA A800 GPUs with 80GB of VRAM. SQL Revision uses Qwen2.5-32B-Instruct as the basic model, and SQL Merge Revision uses Qwen2.5-Coder-32B-Instruct as the basic model.


\begin{figure*}[t]
	\begin{center}
		  \includegraphics[width=1\textwidth]{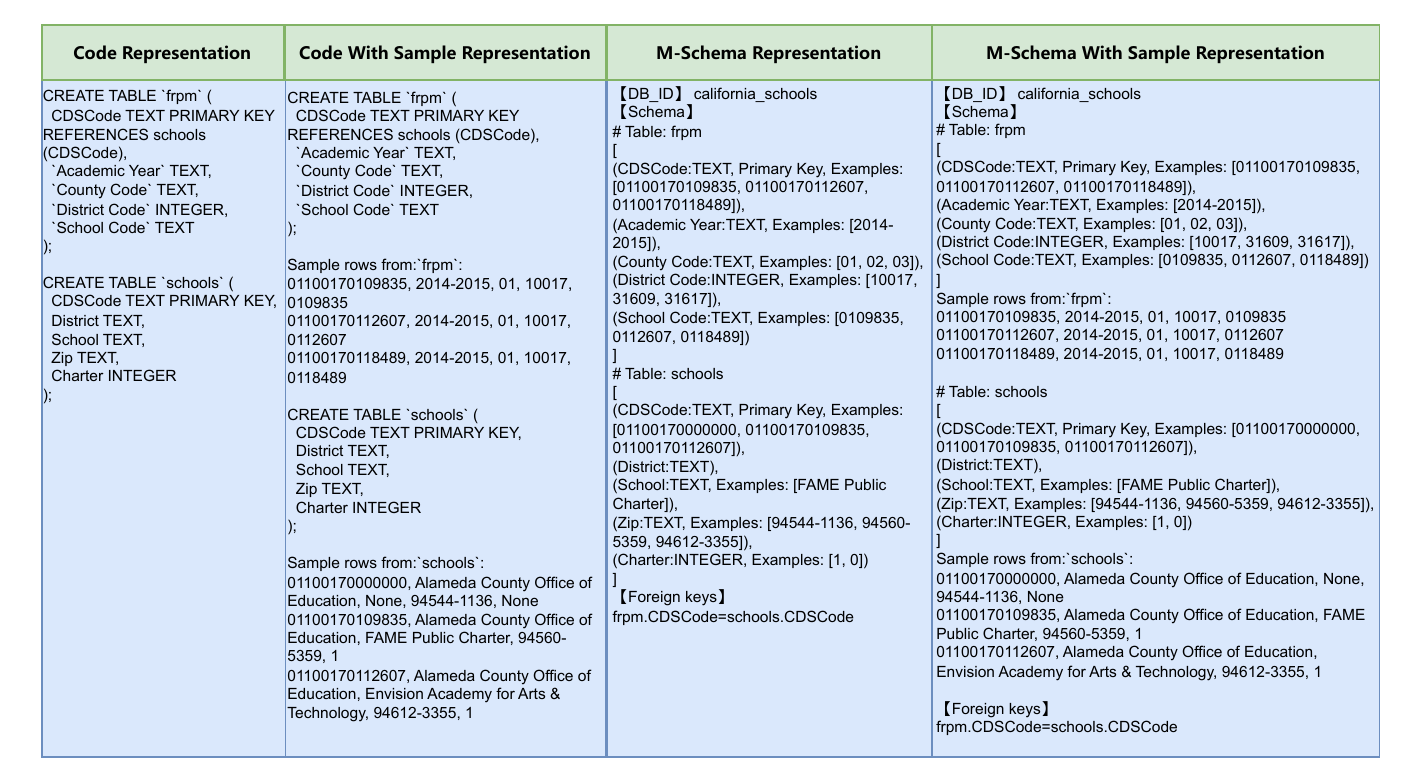}
		  \caption{Examples of database schema representation methods: Code Representation, Code With Sample Representation, M-Schema Representation, and M-Schema With Sample Representation.}
		  \label{figure:schema_representation} 
	\end{center}
\end{figure*}


 \begin{table*}[htbp]
    \centering
	\begin{tabular*}{\textwidth}{llllcc}    
		\toprule
        \text{Method}  & \text{Type}  & \text{Model}   & \text{Size}  & \text{EX(Dev)}  & \text{EX(Test)}\\
		\hline

		\text{XiYan-SQL\citep{xiyansql}}  & \text{\makecell{ICL+\\SFT}}  & \text{GPT-4o} &\text{UNK} & \text{-} & \textbf{89.7}  \\
        \text{MCS-SQL\citep{lee_mcs-sql_2024}}  & \text{ICL}  & \text{GPT-4}  &\text{UNK} & \textbf{89.5} & \text{89.6}  \\
        \text{RSL-SQL\citep{cao2024rsl}}  & \text{ICL}  & \text{GPT-4o}  &\text{UNK} & \text{-} & \text{87.9}  \\
        \text{CHASE-SQL\citep{pourreza2024chase}}  & \text{ICL}  & \text{Gemini 1.5 pro}  & \text{UNK} & \text{-} & \text{87.6}  \\
        \text{CHESS\citep{talaei_chess_2024}}  & \text{ICL}  & \text{Gemini 1.5 pro}  &\text{UNK} & \text{-} & \text{87.2}  \\
        \text{DEA-SQL\citep{xie_decomposition_2024}}  & \text{ICL}  & \text{GPT-4}  &\text{UNK} & \text{85.4} & \text{87.1}  \\
        \text{DAIL-SQL\citep{gao_text--sql_2023}}  & \text{ICL}  & \text{GPT-4}  &\text{UNK} & \text{83.6} & \text{86.6}  \\
        \text{DIN-SQL\citep{pourreza_din-sql_2023}}  & \text{ICL}  & \text{GPT-4}  &\text{UNK} & \text{-} & \text{85.3}  \\
         \midrule
         \text{SENSE\citep{yang_synthesizing_2024}}  & \text{SFT}  & \text{CodeLLaMA}  &\text{13B} & \text{84.1} &  \text{86.6}  \\
         \text{SENSE\citep{yang_synthesizing_2024}}  & \text{SFT}  & \text{CodeLLaMA}  &\text{7B} & \text{83.2} &  \text{83.5}  \\
         \text{MSc-SQL\citep{gorti2024msc}}  & \text{SFT}  & \text{Mistral-v0.3,...}  &\text{9B} & \text{-} & \text{84.7}  \\
         \text{CodeS\citep{li_codes_2024}}  & \text{SFT}  & \text{StarCoder}  &\text{15B} & \text{84.9} & \text{-}  \\
         \text{CodeS\citep{li_codes_2024}}  & \text{SFT}  & \text{StarCoder}  &\text{7B} & \text{85.4} & \text{-}  \\
         \text{DTS-SQL\citep{pourreza-rafiei-2024-dts}}  & \text{SFT}  & \text{DeepSeek}  &\text{7B} & \text{-} &  \text{84.4}  \\
         \text{RESDSQL\citep{li_resdsql_2023}}  & \text{SFT}  & \text{T5}  &\text{3B} & \text{84.1} &  \text{79.9}  \\
         \midrule
         \text{Qwen2.5-Coder\citep{hui2024qwen2}}  & \text{ICL}  & \text{Qwen2.5-Coder}  &\text{32B} & \text{85.1} & \text{85.6}  \\
         \textbf{BASE-SQL(Ours)}  & \text{SFT}  & \text{Qwen2.5-Coder}  &\text{32B} & \text{\underline{86.9}}(\text{$\uparrow$1.8}) & \text{\underline{88.9}}(\text{$\uparrow$3.3})  \\
         \text{Qwen2.5-Coder\citep{hui2024qwen2}}  & \text{ICL}  & \text{Qwen2.5-Coder}  &\text{14B} & \text{84.1} & \text{85.1}  \\
         \textbf{BASE-SQL(Ours)}  & \text{SFT}  & \text{Qwen2.5-Coder}  &\text{14B} & \text{\underline{86.8}}(\text{$\uparrow$2.5}) & \text{\underline{87.9}}(\text{$\uparrow$2.8})  \\
		\bottomrule
	\end{tabular*}
	\caption{Performance Comparison of different Text-to-SQL methods on Spider dev and test dataset.}
	\label{tab:result_main_spider}
	\footnotetext{}
\end{table*}

\section{Analysis}
\label{sec:appendix_analysis}

\subsection{Impacts of Different Schema Representations}
\label{subsec:impacts_of_different_schema_representations}

\begin{table*}[htbp]
    \centering
	\begin{tabular}{l|cc|cc|cc}    
		\toprule
        \text{Method} & \text{32B}&  \text{SFT-32B}  & \text{14B}&  \text{SFT-14B} & \text{7B}&  \text{SFT-7B}\\
        \midrule
        \text{Full Code}  & \text{56.84} & \text{60.63}   & \text{56.45} & \text{55.67} & \text{49.48} & \text{41.40} \\
        \text{Full M-Schema}  & \text{58.21} & \text{61.73}   & \text{56.39} & \text{60.04} & \text{50.33} & \text{41.79}\\
        \text{Full Code + Sample}  & \text{59.39} & \textbf{62.78}   & \text{56.71} & \text{58.74} & \textbf{50.91} & \text{44.65}\\
        \text{Full M-Schema + Sample}  & \textbf{59.71} & \text{62.19}   & \textbf{57.50} & \textbf{60.82} & \text{50.59} & \textbf{45.31}\\
        \midrule
        \text{Schema Linking Code}  & \text{56.78} & \text{61.93}   & \text{55.67} & \text{58.87} & \text{51.30} & \text{55.48}\\
        \text{Schema Linking M-Schema}  & \text{59.71} & \text{63.56}   & \text{57.50} & \text{61.34} & \text{53.46} & \text{58.41}\\
        \text{Schema Linking Code + Sample}  & \text{59.65} & \text{63.49}   & \textbf{60.43} & \text{61.93} & \text{53.72} & \text{57.82}\\
        \text{Schema Linking M-Schema + Sample}  & \textbf{61.54} & \textbf{64.21}   & \text{58.08} & \textbf{62.32} & \textbf{54.95} & \textbf{58.74}\\
		\bottomrule
	\end{tabular}
	\caption{Execution accuracy(EX) of different schema representations on the BIRD development set.}
	\label{tab:result_different_schema}
	\footnotetext{}
\end{table*}

In order to show the impact of different schema representations on SQL generation, we used full-schema to conduct experiments on the BIRD development set. The experimental results of three different sizes of Qwen2.5-Coder and its supervised fine-tuning are shown in Table \ref{tab:result_different_schema}. Through the experimental results, we found that the M-Schema representation is more advantageous than the Code representation after adding different sample values of the column, and the performance of SQL generation can be further improved after adding table content. Whether before or after fine-tuning, the M-Schema with sample representation has the best performance.

\subsection{Impacts of Candidate Generation}
\label{subsec:impacts_of_candidate_generation}

\begin{figure*}
	\begin{center}
		  \includegraphics[width=1\textwidth]{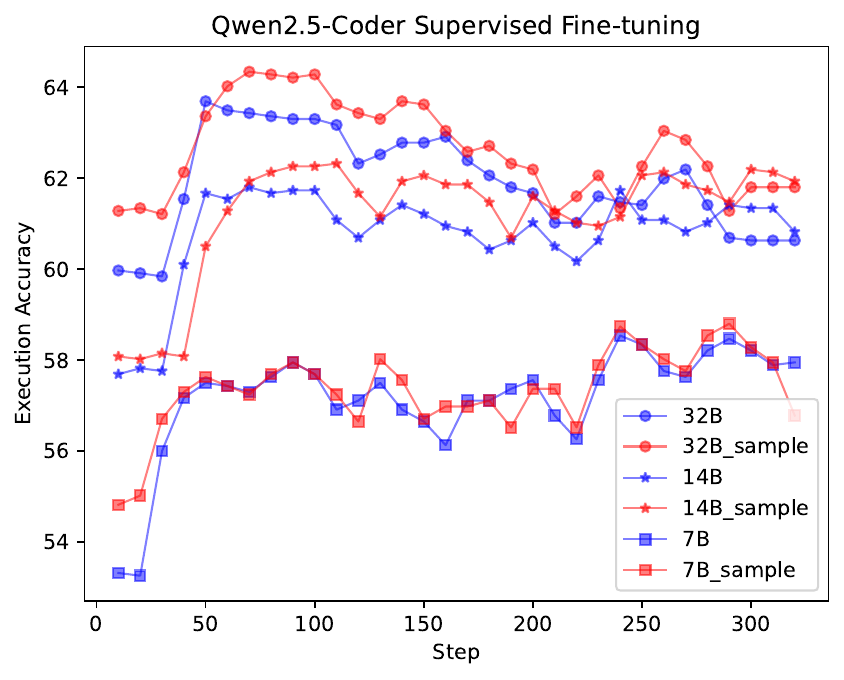}
		  \caption{After using Qwen2.5-Coder of different sizes for Supervised Fine-tuning, the execution accuracy on the BIRD development set after evaluating different steps. The batch size of the 32B and 14B models is 8, and the batch size of the 7B model is 16. In the legend, 32B, 14B, and 7B use the M-Schema representation, while 32B\_{}sample, 14B\_{}sample, and 7B\_{}sample use the M-Schema with sample representation.}
		  \label{figure:sft_step} 
	\end{center}
\end{figure*}

In Figure \ref{figure:sft_step}, we use models of different sizes for fine-tuning, showing the relationship between the number of fine-tuning steps and EX. We can see that for models of the same size, the M-Schema with sample representation is better than the M-Schema representation. The 32B and 14B models tend to stabilize after about 150 steps (about 1.2k samples are trained), and the 7B model tends to stabilize after about 250 steps (about 4k samples), which shows that Qwen2.5-Coder has a strong generalization ability and can achieve good results by fine-tuning only on a small number of samples.

\subsection{Impacts of Different SQL Section Methods}

\begin{table}[htbp]
	\begin{tabular}{lcc}    
		\toprule
        \text{Method} & \text{\makecell{BIRD\\EX(Dev)}}  \\
        \midrule
        \text{src}  & \text{65.78 + 64.28}  \\
        \text{upper bound}  & \text{68.25}  \\
        \midrule
        \text{XiYan-SQL Selection Prompt}  & \text{66.62}  \\
        \text{MSc-SQL Selection Prompt}  & \text{66.56}  \\
        \text{CHASE-SQL Selection Prompt}  & \text{66.43}  \\
        \text{Our Merge Revision}  & \textbf{66.95}   \\
		\bottomrule
	\end{tabular}
	\caption{Execution accuracy(EX) of different SQL selection methods on the BIRD development set.}
	\label{tab:result_different_sql_selection}
	\footnotetext{}
\end{table}

Recently, the model-based SQL selection method has received attention and achieved good results \citep{xiyansql,gorti2024msc, pourreza2024chase}. Therefore, we selected two candidate SQLs and then used Qwen2.5-Coder-32B-Instruct model to conduct comparative experiments with different SQL selection methods. The results are shown in Table \ref{tab:result_different_sql_selection}. It can be seen from the table that our merge revision method is better than the method based on Selection Prompt.

In order to show the intermediate process of our merge revision method in detail, we used different initialization settings to conduct SQL revision experiments. The experimental results are shown in Table \ref{tab:result_different_sql_revision_detail}. From the results, we can find that candidate SQL1 can get the best performance by only performing one round of sql revision and three merge revision (SQL5, SQL6, SQL8). Merge revision is only required when the results of the two candidate SQLs are inconsistent after execution. If the execution results are consistent, the first SQL is directly output. Therefore, there are not many cases where merge revision is really needed.

\begin{table*}[htbp]
    \centering
	\begin{tabular}{l|cc|cc}    
		\toprule
        \text{Method} & \text{\makecell{1-time\\revision}} & \text{\makecell{2-time\\revision}}  & \text{\makecell{1-time\\revision}}  & \text{\makecell{2-time\\revision}} \\
        \midrule
        \text{SQL1}  & \text{64.21}   & \text{64.21} & \text{64.15} & \text{64.15}  \\
        \text{SQL2(Full Schema)}  & \text{66.04}   & \text{66.30} & \text{65.91}   & \text{66.23}\\
        \text{SQL3(Full With Sample Schema)}  & \text{65.91}    & \text{65.84} & \text{65.58}   & \text{65.58}\\
        \midrule
        \text{SQL4(SQL1+SQL2)}  & \text{66.69}   & \text{66.95}  & \text{66.43} & \text{66.69} \\
        \text{SQL5(SQL2+SQL3)}  & \text{67.14}  & \text{66.82}  & \text{66.88}  & \text{66.62}\\
        \text{SQL6(SQL1+SQL3)}  & \text{67.01}   & \text{66.95} & \text{66.75}  & \text{66.62}\\
        \midrule
        \text{SQL7(SQL4+SQL5)}  & \text{67.34}  & \textbf{67.41} & \text{67.14}   & \textbf{66.95}\\
        \text{SQL8(SQL5+SQL6)}  & \textbf{67.47}   & \text{67.14} & \textbf{67.21}  & \textbf{66.95}\\
        \text{SQL9(SQL4+SQL6)}  & \text{67.14}   & \text{67.14}  & \text{66.95} & \text{66.75}\\
		\bottomrule
	\end{tabular}
	\caption{Detailed evaluation results of SQL revision and SQL merge revision using different initial candidate SQL1 on the BIRD development set. 1-time revision means that only one round of SQL revision is performed on candidate SQL1 to generate SQL2 and SQL3, and 2-time revision means that two rounds of SQL revision are performed on candidate SQL1 to generate SQL2 and SQL3. SQL4 (SQL1+SQL2) means that candidate SQL4 is generated after merge revision of SQL1 and SQL2. The generation process of other SQL5-SQL9 is similar.}
	\label{tab:result_different_sql_revision_detail}
	\footnotetext{}
\end{table*}

\section{Prompt Templates}
\label{sec:appendix_prompt}

\subsection{Prompt for Schema Linking}
\label{sec:prompt_schema_linking}
\noindent
\begin{minipage}{\columnwidth}
    \begin{mdframed}
    \{DATABASE SCHEMA\}\\
    \\
    \mbox{-}\mbox{-} Given the previous table schema combined with the additional information provided, help me find all the table names associated with answering the user's question.\\
    \\
    Question: \{QUESTION\}\\
    \mbox{-}\mbox{-} External Knowledge: \{EVIDENCE\}\\
    
    Remember not to generate SQL, but reply with the relevant table names. Please reply in JSON format:\\
    ```json\\
    \{\\
        "tables": ["table1","table2",..]\\
    \}\\
    ```\\
    \end{mdframed}
    \label{fig:prompt_schema_linking}
\end{minipage}

\subsection{Prompt for Candidate SQL Generate}
\label{sec:prompt_candidate_sql_generate}
\noindent
\begin{minipage}{\columnwidth}
    \begin{mdframed}
    \{DATABASE SCHEMA\}\\
    \\
    \mbox{-}\mbox{-} Using valid SQLite and understanding External Knowledge, answer the following questions for the tables provided above.\\
    \\
    Question: \{QUESTION\}\\
    \mbox{-}\mbox{-} External Knowledge: \{EVIDENCE\}\\
    
    Please output only the final SQL query, starts with keyword `SELECT`.
    \end{mdframed}
    \label{fig:prompt_candidate_sql_generate}
\end{minipage}

\subsection{Prompt for SQL Revision}
\label{sec:prompt_sql_revision}
\noindent
\begin{minipage}{\columnwidth}
    \begin{mdframed}
    \{DATABASE SCHEMA\}\\
    \\
    \mbox{-}\mbox{-} Using valid SQLite and understanding External Knowledge, revise the SQL query that answers the following questions of the above table schema based on the predicted SQL and SQL execution results. If the current SQL query is correct, return the query directly.\\
    \\
    Question: \{QUESTION\}\\
    \mbox{-}\mbox{-} External Knowledge: \{EVIDENCE\}\\
    
    Predicted SQL query: \{PREDICT\_SQL\} \\
    SQL execute result: \{EXECUTE\_RESULT\}\\

    Please output only the final revised SQL query, starts with keyword `SELECT`.
    \end{mdframed}
    \label{fig:prompt_sql_revision}
\end{minipage}

\subsection{Prompt for SQL Merge Revision}
\label{sec:prompt_sql_merge_revision}
\noindent
\begin{minipage}{\columnwidth}
    \begin{mdframed}
    \{DATABASE SCHEMA\}\\
    \\
    \mbox{-}\mbox{-} Using valid SQLite and understanding External Knowledge, answer the following questions for the tables provided above. \\
    \\
    Question: \{QUESTION\}\\
    \mbox{-}\mbox{-} External Knowledge: \{EVIDENCE\}\\
    
    Here are some corresponding draft SQL and execute result:\\
    1. \{PREDICT\_SQL1\} \\
    Execution result\\
    \{EXECUTE\_RESULT1\}\\

    2. \{PREDICT\_SQL2\} \\
    Execution result\\
    \{EXECUTE\_RESULT2\}\\

    Please output only the final SQL query, starts with keyword `SELECT`.
    \end{mdframed}
    \label{fig:prompt_sql_merge_revision}
\end{minipage}

\end{document}